\documentclass{article}
\usepackage{cite}
\usepackage{url}
\usepackage{amsmath,amssymb,amsfonts}
\usepackage{algorithmic}
\usepackage{graphicx}
\usepackage{epsfig}
\usepackage{textcomp}
\usepackage{xcolor}
\usepackage{textpos}

\def\BibTeX{{\rm B\kern-.05em{\sc i\kern-.025em b}\kern-.08em
    T\kern-.1667em\lower.7ex\hbox{E}\kern-.125emX}}
\begin{document}

\title{Functional Specification of the RAVENS Neuroprocessor}

\author{Adam Z. Foshie \and James S. Plank \and Garrett S. Rose \and Catherine D. Schuman \\
\mbox{}\\
EECS Department \\
University of Tennessee \\
Knoxville, TN 37996 \\
{\tt [afoshie,jplank,garose,cschuman]@utk.edu }}

\maketitle

\begin{abstract}
RAVENS is a neuroprocessor that has been developed by the TENNLab research
group at the University of Tennessee.  Its main focus has been as a vehicle
for chip design with memristive elements; however it has also been the vehicle
for all-digital CMOS development, plus it has implementations on FPGA's,
microcontrollers and software simulation.  The software simulation is supported
by the TENNLab neuromorphic software framework so that researchers may develop
RAVENS solutions for a variety of neuromorphic computing applications.
This document provides a functional specification of RAVENS that should apply 
to all implementations of the RAVENS neuroprocessor.
\end{abstract}


\section{Introduction}

Neuromorphic computing is an active research area, with one of the key activities being
the design of brain-based neuroprocessors.  Examples are SpiNNaker~\cite{fgt:14:tsp},
TrueNorth~\cite{a:15:tdt-long} and Loihi~\cite{d:18:loihi-long}.  The TENNLab research
group has designed multiple neuroprocessors and implemented them in/on CMOS, FPGA's 
and/or microcontrollers.  This activity started
with DANNA~\cite{drc:16:dns},
and includes DANNA2~\cite{mdb:18:dda}, mrDANNA~\cite{cdr:16:hom}, Caspian~\cite{msp:20:cnd}
and RISP~\cite{pzg:22:risp}.
Informed by past experience, we have designed the RAVENS (Reconfigurable and Very Efficient Neuromorphic System) neuroprocessor architecture.  This neuroprocessor
has implementations in CPU simulation, microcontroller, and FPGA, with CMOS and mixed CMOS/memristor
implementations in development.  

This document provides a functional specification of RAVENS, which applies to all implementations.
The RAVENS simulator is part of the TENNLab software framework for neuromorphic 
computing~\cite{spp:21:sfc,psd:18:ten}, and as such may be trained for a variety of
applications.  Its FPGA and microcontroller implementations may be used as the
neuroprocessors that are central to the Neuromorphic Starter Kit
for embedded computing applications~\cite{pgf:22:dns}.

RAVENS is a processor for spiking neural networks.  
The field of spiking neural networks has had a long history.  The reader is
referred to articles by Roy {\em et al}~\cite{rjp:19:twb} and Schuman {\em et al}~\cite{spp:17:sur}
for historical overviews of spiking neural networks and their hardware implementations.

\section{Hardware Constants and Network Settings}

In all RAVENS implementations, there are parameters that must be set when the neuroprocessor
is created, and remain constant for the implementation.  We call these {\em Hardware
Constants}.  For example, one such parameter is
the maximum delay in a synapse.  One typically sets these parameters using a
JSON specification file or an HDL package file, depending on the implementation platform. Then the software that creates the implementation does so using
the values in the relevant file.

For example, when RAVENS is being implemented on FPGA, the maximum synapse delay
is read from the HDL package file, and then the neuroprocessor implementation of the FPGA instantiates registers of this
width such that it will not change.  These values have implications on all facets of the
neuroprocessor.  For example, larger values of maximum synapse delay consume more processing
resources than smaller values, and can result in some of the following implications when
implemented:

\begin{itemize}
\item Fewer neurons and synapses being implemented on the neuroprocessor.
\item Longer processing times.
\item Increased power consumption.
\end{itemize}

Any implementation of RAVENS should be accompanied by a software module that reports the
implications of the Hardware Constants.  Obviously, the RAVENS software simulator is more flexible
with respect to these constants.

{\em Network Settings} are values that are set by the neural network which is loaded onto
the neuroprocessor.  These settings are partitioned into {\em Neuron Settings}, which apply
to individual neurons, {\em Synapse Settings}, which apply to individual synapses, and
{\em Overall Settings}, which apply to the network as a whole.  The Network Settings are
defined by a JSON file, typically created with the TENNLab software framework~\cite{spp:21:sfc,psd:18:ten}. All settings are signed or unsigned integers.

\section{Neurons and nCores}

At a high level, RAVENS is composed of neurons and synapses like most spiking neural networks.
There is an additional concept of a neural core, or {\em nCore}, each of which contains multiple neurons and
synaptic connections.  Some of the Hardware Constants apply to nCores rather than neurons and
synapses. 
In the presentation below, we will at times refer to a neuron's ``charge,'' which is equivalent
to its action potential, and a synapse's ``charge,'' which is equivalent to its weight.

\subsection{Neuron Characteristics and Settings}

Each neuron maintains an {\em activation potential} over time.  This is a discrete value
stored in an accumulation register or other memory component.  
For example, an analog implementation may store the activation potential as an analog voltage across a capacitor in an integrate-and-fire neuron.  Here, we generally refer to these storage elements as registers for the sake of simplicity.  
The register value can be represented as a signed integer whose number
of bits is a Hardware Constant.  Its value starts at a {\em standard resting potential},
which is a Neuron Setting.  The value may be increased or decreased by spikes on incoming
synapses.  The magnitude of increase is equal to the synapse's weight (a Synapse Setting).
When the activation potential exceeds the neuron's {\em threshold} (another Neuron Setting),
then the neuron will fire, which will place spikes on all of its outgoing synapses.  The threshold
is a signed integer whose number of bits are a hardware constant.

Neurons accumulate and fire on discrete integration cycles.  
At the beginning of the cycle, the accumulation register's value is compared to the threshold,
and if it exceeds the threshold, the neuron fires.  At that point, charge is put into the 
neuron's post-synapses and that charge is sent to the synapses' post-neurons.  The neuron
may then enter one of the two refractory periods described below.  When the refractory
periods are over, then the neuron enters {\em standard operation}, where charge is
added to or subtracted from the neuron's accumulation register.  There is a Neuron Setting
called the {\em standard resting potential}.  When the neuron enters standard operation,
if its threshold is less than its standard resting potential, then it is set to the standard
resting potential.  While in standard operation, the neuron may not end its integration
with its accumulation register less than its standard resting potential.  If the value of
the register is less than the standard resting potential at the end of the integration cycle,
then it is set to the standard resting potential.

We note that a neuron may end its integration cycle with its accumulation register having
a value above the neuron's threshold.  In that case, the neuron will spike at the beginning
of the next integration cycle.

{\bf Refractory Periods.} Neurons have two refractory periods after they spike.  These
are Neuron Settings.  The first is an {\em absolute refractory period}.  Its minimum value
is zero and its maximum value is a Hardware Constant. It is specified in cycles, and for
the given number of cycles after a neuron spikes, it will not accumulate spikes from
incoming synapses.  

When the absolute refractory period ends, there is a {\em relative refractory
period,} which is another Neuron Setting, whose minimum value is zero and whose
maximum value is a Hardware Constant.  At the beginning of this refractory
period, the activation potential is set to a {\em refractory resting
potential}, which is another Neuron Setting.  During this refractory period,
spikes are processed normally, but the shift in resting potential adjusts the
effective threshold for the duration of the relative refractory period.  Typically,
the refracting resting potential is lower than the standard resting potential,
which means that more spiking stimulus is required to produce a spike during this
period.  Like the standard operation, during the relative refractory
period, the accumulation register may not end an integration cycle with its value
below the refractory resting potential.  If its value is below the refractory
resting potential, then it is set to the refractory resting potential.
At the end of the
refractory period, if the activation potential is less than the standard
resting potential, then it is set to the standard resting potential so that its
regular operation is not inhibited.

{\bf Leak.} Leak is a Neuron Setting whose minimum value is zero and whose maximum
value is a Hardware Constant. During standard operation,
leak works as follows: If, at the start of a cycle, 
the action potential's value is above the
standard resting potential, then the value of the leak is subtracted from the action potential.
If the result is smaller than the standard resting potential, then it is set to 
standard resting potential.

During the absolute refractory period, leak is ignored.  During the relative refractory
period, leak is applied to the action potential only if its value is greater than the refractory
resting potential.
If the result is smaller than the relative resting potential, then it is set to the 
refractory resting potential. 

{\bf Ports.} Each neuron has a fixed number of ``ports'', which is a Hardware Constant.
The ports may be employed either for pre-synapses or for charge injection, which is described
below under {\em Synapses}.

\subsection{Synapse Characteristics and Settings}

As with all spiking neuroprocessors, RAVENS synapses have delays and weights.
Each is a Synapse Setting. Delays are non-negative numbers whose maximum value is a 
Hardware Constant. Weights are signed integers whose number of bits is a Hardware Constant.
When a neuron fires at the beginning of an integration cycle, its synapses start to communicate
their spikes.  The synapse weights are delivered to their post-neurons on the~$d-th$ integration
cycle after the fire, where~$d$ is the synapse's delay.  We note that delays of zero are
possible, which means that a synapse's weight is delivered to its post neuron on the same 
integration cycle as the pre neuron's firing.  This is possible because neurons fire at the
beginning of the integration cycle.

Each neuron in a RAVENS network has a maximum number of synapses that may be attached to it.
This is a Hardware Setting.

{\bf STDP}: Synapse weights may be modified at run time depending on
the relationship between the last synaptic fire time and the
last neuron fire time.  This modification is specified by
a lookup table called the STDP table.  The table is a Hardware Constant.
Potentiation and depression work in different ways, but they use the 
same table.  We describe them separately.

{\em Potentiation.}  When a neuron's action potential exceeds its threshold at the end
of an integration cycle, it will fire at the beginning of the next cycle.  
Before that
next cycle begins, each of the neuron's pre-synapses is checked to see
when it last fired into the neuron.  Suppose the potential exceeds its threshold at the
end of integration cycle~$y$,
and one of its pre-synapses last fired into it at integration cycle~$x$.  We turn
these numbers into a table index with the following equation:

\begin{equation}
    index = \left\lfloor\frac{T}{2}\right\rfloor - (y-x),
\end{equation}

where~$T$ is the size of the STDP table.  If $index \ge 0$, 
then we use the value in that index of the STDP table to increase the
synapse's weight (up to its maximum weight).  Here is an example.  Suppose that the
STDP table is [ 1, 2, 2, 3, 4, -4, -2, -1.  And suppose that a neuron $n$ has three
pre-synapses, $a$, $b$ and~$c$.  Finally, suppose the following events occur:

\begin{itemize}
\item At timestep 7, $a$ fires into $n$, incrementing its action potential, but not so that
      it exceeds its threshold.
\item At timestep 10, $b$ fires into $n$, incrementing its action potential, but again, not so that
      it exceeds its threshold.
\item At timestep 12, $c$ fires into $n$, incrementing its action potential so that it does
      exceed its threshold.
\item At the beginning of timestep 13, $n$ will fire.  
\end{itemize}

At the end of timestep 12, synapses~$a$, $b$ and $c$ are checked for potentiation.  Their
indices will be -1, 2 and 4 respectively, meaning synapse~$b$'s weight will potentiate by 2,
and synapse~$c$'s weight will potentiate by 4.

For depression, at the end of each integration cycle, if a neuron received charge from a 
synapse, but its potential does not exceed its threshold, then the synapse is a candidate
for depression.  Suppose we are at cycle~$y$ and the neuron last fired at the beginning
of cycle~$x$.  Then we calculate a table index with the following equation:

\begin{equation}
    index = \left\lfloor\frac{T}{2}\right\rfloor + (y-x),
\end{equation}

If the index is a valid index of the STDP table, then its value is used to decrement the
synapse's weight.

{\bf Charge Injection}: There are two ways to apply external input to a RAVENS network.
The first is to simply cause one of the neuron's pre-synapses to spike.  The second way is
to allocate a number of synapse ``ports'' for {\em charge injection}.  This number is a 
Hardware Constant.  Then, it is a Neuron Setting to enable the ports.  If a neuron's ports
are enabled, then input values may be delivered to the neuron using the ports as a signed
binary number.

\subsection{Minimum Accumulation Register Width}

The large amount of configurability regarding the resolution of synapse weights, the use of charge injection, and the number of synapse ports means that some care must be taken when implementing the charge accumulation registers. If these registers are made too small, inputs can inadvertently cause overflows to occur in the calculation of the total charge before a proper comparison is made. To ensure that the width of the accumulation registers is sufficient for any programmed scenario given a set of Hardware Constants, the minimum accumulator width is set by the following equation: 

\begin{equation}
  A = \lceil \log_2 \left[ \max 
  \left\{ 
  \begin{matrix} 
    \left([2^W-1] \cdot [S-C] + 2^C-1\right) \\ 
    \left([2^W-1] \cdot S \right) 
  \end{matrix} 
  \right. 
  \right]\rceil 
\end{equation}

Where $A$ is the accumulator width, $W$ is the width of the synapse weights, $S$  is the number of synapses, and $C$  is the number of synapse ports used for charge injection when enabled.

\section{Examples}

To help illustrate the descriptions above, in the following subsections, we show example networks and how they run on RAVENS.  In this presentation, the word ``timestep'' is equivalent
to ``integration cycle.''

\subsection{Simple integrate and fire}

\begin{figure}[ht]
\begin{center}
\includegraphics[scale=0.80]{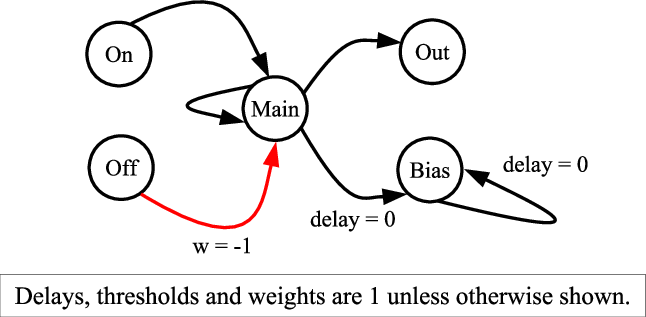}
\caption{\label{fig:network_1_basic} Simple integrate-and-fire network.}
\end{center}
\end{figure}

We start with the network pictured in Figure~\ref{fig:network_1_basic}, and the following
RISP settings: 

\begin{itemize}
\item No absolute refractory period.
\item No relative refractory period.
\item No leak.
\item No STDP.
\item Standard resting potential is zero.
\item The neurons {\em Main}, {\em On} and {\em Off} have synapses with weights of 16
      that may be fired from the external environment.  
      When we say we ``apply input'' to 
      one of these neurons, what we mean is that we cause the appropriate one of these
      synapses to fire, which add 16 to the neuron's potential, and 
      causes the neuron to fire at the beginning of
      the next timestep.
\end{itemize}

In Table~\ref{tab:network_1_basic}, we show the activity of the network over 15 timesteps,
when we apply input to $Main$ at timestep 0, $On$ at timestep 5, and $Of\/f$ at timestep 10.
It is important to note that neurons spike at the beginning of the timestep, so when charge
is applied at timestep~$t$, that raises the appropriate neuron's charge to 16 at the end
of the timestep.  This causes the neuron to fire at the beginning of the next timestep.
This is why $Main$, $On$ and $Of\/f$ fire at timesteps 1, 6 and 11 respectively.

Next, if a synapse's delay is $d$, and its pre-neuron fires at timestep~$t$, then it will
fire into its post-neuron at timestep~$t+d$.  For example, the synapse from $Main$ to $Bias$
has a delay of 0, so it fires into $Bias$ at timestep 1, increasing $Bias'$ charge to 1.
The synapse from $Main$ to $Out$ has a delay of 1, so it fires into $Out$ at timestep 2.

Finally, neurons only spike when their thresholds are exceeded.  This is why, for example,
$Bias$ ends timestep 1 with a charge of 1, but it does not spike at the beginning of
timestep 2.  When the spike from $On$ arrives at $Main$ during timestep 7, $Main's$ charge
is raised to 2, and accordingly it fires at timestep 8.

The inhibitory synapse from $Of\/f$ to $Main$ fires at timestep 11, decreasing $Main's$
charge to zero at timestep 12.

\begin{table}[ht]
\begin{center}
\begin{tabular}{|l|c|ccccc|ccccc|l|}
\hline
Timestep & Apply & \multicolumn{5}{c|}{Fire} &
                   \multicolumn{5}{c|}{Charge at end of timestep} \\
         & spike to & $Main$ & $On$ & $Of\/f$ & $Out$ & $Bias$ &
                       $Main$ & $On$ & $Of\/f$ & $Out$ & $Bias$ \\
\hline
  0 & $Main$ & - & - & - & - & - & 16 & 0 & 0 & 0 & 0  \\
  1 &   -    & * & - & - & - & - & 0 & 0 & 0 & 0 & 1  \\
  2 &   -    & - & - & - & - & - & 1 & 0 & 0 & 1 & 1  \\
  3 &   -    & - & - & - & - & - & 1 & 0 & 0 & 1 & 1  \\
  4 &   -    & - & - & - & - & - & 1 & 0 & 0 & 1 & 1  \\
  5 & $On$   & - & - & - & - & - & 1 & 16 & 0 & 1 & 1  \\
  6 &   -    & - & * & - & - & - & 1 & 0 & 0 & 1 & 1  \\
  7 &   -    & - & - & - & - & - & 2 & 0 & 0 & 1 & 1  \\
  8 &   -    & * & - & - & - & - & 0 & 0 & 0 & 1 & 2  \\
  9 &   -    & - & - & - & - & * & 1 & 0 & 0 & 2 & 1  \\
 10 & $Of\/f$  & - & - & - & * & - & 1 & 0 & 16 & 0 & 1  \\
 11 &   -    & - & - & * & - & - & 1 & 0 & 0 & 0 & 1  \\
 12 &   -    & - & - & - & - & - & 0 & 0 & 0 & 0 & 1  \\
 13 &   -    & - & - & - & - & - & 0 & 0 & 0 & 0 & 1  \\
 14 &   -    & - & - & - & - & - & 0 & 0 & 0 & 0 & 1  \\
\hline
\end{tabular}
\caption{\label{tab:network_1_basic} Activity of the network in Figure~\ref{fig:network_1_basic}.}
\end{center}
\end{table}

\begin{figure}[ht]
\begin{center}
\includegraphics[scale=0.80]{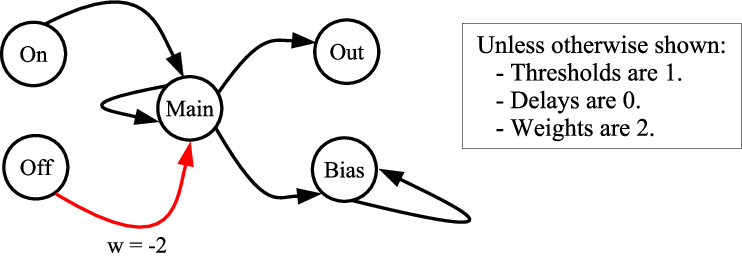}
\caption{\label{fig:network_2_every_timestep} A network set so that $Main$ fires every
timestep when turned ``on.''}
\end{center}
\end{figure}

The network shown in Figure~\ref{fig:network_2_every_timestep} is similar to the
network in Figure~\ref{fig:network_1_basic}, except the synapse weights have been
multiplied by two, and the synapse delays have all been set to zero.  The point of
this network is to make $Main$ fire every timestep when it is turned ``on.'' You can 
turn $Main$ on and off by applying a spike to the $On$ and $Of\/f$ neurons.  Whenever $Main$
spikes, $Out$ spikes one timestep later, and when $Main$ spikes for the first time,
$Bias$ will start spiking one timestep later.  $Bias$ will then continue to spike
every timestep, regardless of what happens to $Main$.

In Table~\ref{tab:network_2_every_timestep}, we apply spikes to $On$ at timesteps 0 and 8,
and spikes to $Of\/f$ at timesteps 4 and 12.  This causes $Main$ to start spiking at timesteps
2 and 10, and to stop spiking at timesteps 6 and 14:

\begin{table}[ht]
\begin{center}
\begin{tabular}{|l|c|ccccc|ccccc|l|}
\hline
Timestep & Apply & \multicolumn{5}{c|}{Fire} &
                   \multicolumn{5}{c|}{Charge at end of timestep} \\
         & spike to & $Main$ & $On$ & $Of\/f$ & $Out$ & $Bias$ &
                       $Main$ & $On$ & $Of\/f$ & $Out$ & $Bias$ \\
\hline
  0 & $On$   & - & - & - & - & - & 0 & 16 & 0 & 0 & 0  \\
  1 &    -   & - & * & - & - & - & 2 & 0 & 0 & 0 & 0  \\
  2 &    -   & * & - & - & - & - & 2 & 0 & 0 & 2 & 2  \\
  3 &    -   & * & - & - & * & * & 2 & 0 & 0 & 2 & 4  \\
  4 & $Of\/f$  & * & - & - & * & * & 2 & 0 & 16 & 2 & 4  \\
  5 &    -   & * & - & * & * & * & 0 & 0 & 0 & 2 & 4  \\
  6 &    -   & - & - & - & * & * & 0 & 0 & 0 & 0 & 2  \\
  7 &    -   & - & - & - & - & * & 0 & 0 & 0 & 0 & 2  \\
  8 & $On$   & - & - & - & - & * & 0 & 16 & 0 & 0 & 2  \\
  9 &    -   & - & * & - & - & * & 2 & 0 & 0 & 0 & 2  \\
 10 &    -   & * & - & - & - & * & 2 & 0 & 0 & 2 & 4  \\
 11 &    -   & * & - & - & * & * & 2 & 0 & 0 & 2 & 4  \\
 12 & $Of\/f$  & * & - & - & * & * & 2 & 0 & 16 & 2 & 4  \\
 13 &    -   & * & - & * & * & * & 0 & 0 & 0 & 2 & 4  \\
 14 &    -   & - & - & - & * & * & 0 & 0 & 0 & 0 & 2  \\
 15 &    -   & - & - & - & - & * & 0 & 0 & 0 & 0 & 2  \\
\hline
\end{tabular}
\caption{\label{tab:network_2_every_timestep} Activity of the network in Figure~\ref{fig:network_2_every_timestep}.}
\end{center}
\end{table}

\clearpage

\subsection{Leak and Standard Resting Potential}

\begin{figure}[ht]
\begin{center}
\includegraphics[scale=0.80]{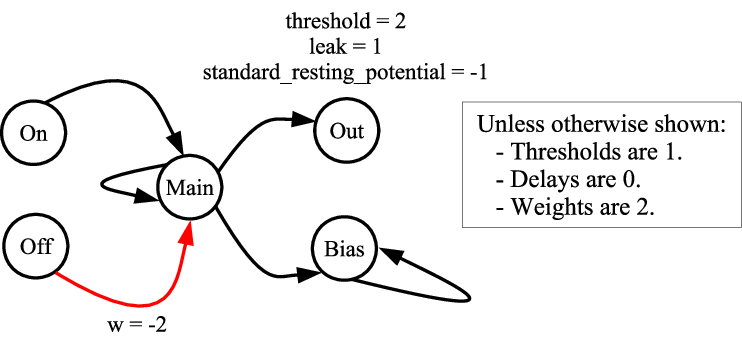}
\caption{\label{fig:network_3_leak} A network to demonstrate leak and the standard resting
otential.}
\end{center}
\end{figure}

We next test the network depicted in Figure~\ref{fig:network_3_leak}.
This network configures $Out$ to so that its threshold is 2, its leak is 1 and
its standard resting potential is -1.  

In Table~\ref{tab:network_3_leak}, we show what happens when we apply a spike to $Main$
at timestep 0.  $Main$ spikes every timestep, adding two units of charge to $Out$.
$Out$ starts with its potential equal to its standard resting potential, which is -1.
At timestep 1, the synapse from $Main$ to $Out$ increases $Out's$ potential to 1.
At the beginning of timestep 2, the charge leaks from 1 to 0, and then the synapse
from $Main$ raises the charge from 0 to 2.  
Similarly, at the beginning of timestep 3, the charge leaks from 2 to 1, and then the synapse
from $Main$ raises the charge from 0 to 3.   That means $Out$ will fire at timestep 4,
which it does.  

After firing, $Out$'s potential is reset to -1, and then the charge from $Main$ increments
it to 1.  The pattern continues, which is why $Out$ fires every three timesteps.

\begin{table}[ht]
\begin{center}
\begin{tabular}{|l|c|ccccc|ccccc|l|}
\hline
Timestep & Apply & \multicolumn{5}{c|}{Fire} &
                   \multicolumn{5}{c|}{Charge at end of timestep} \\
         & spike to & $Main$ & $On$ & $Off$ & $Out$ & $Bias$ &
                       $Main$ & $On$ & $Off$ & $Out$ & $Bias$ \\
\hline
  0 & $Main$ & - & - & - & - & - & 16 & 0 & 0 & -1 & 0  \\
  1 &    -   & * & - & - & - & - & 2 & 0 & 0 & 1 & 2  \\
  2 &    -   & * & - & - & - & * & 2 & 0 & 0 & 2 & 4  \\
  3 &    -   & * & - & - & - & * & 2 & 0 & 0 & 3 & 4  \\
  4 &    -   & * & - & - & * & * & 2 & 0 & 0 & 1 & 4  \\
  5 &    -   & * & - & - & - & * & 2 & 0 & 0 & 2 & 4  \\
  6 &    -   & * & - & - & - & * & 2 & 0 & 0 & 3 & 4  \\
  7 &    -   & * & - & - & * & * & 2 & 0 & 0 & 1 & 4  \\
  8 &    -   & * & - & - & - & * & 2 & 0 & 0 & 2 & 4  \\
  9 &    -   & * & - & - & - & * & 2 & 0 & 0 & 3 & 4  \\
 10 &    -   & * & - & - & * & * & 2 & 0 & 0 & 1 & 4  \\
\hline
\end{tabular}
\caption{\label{tab:network_3_leak} Activity of the network in Figure~\ref{fig:network_3_leak}.
The leak causes $Out$ to fire every three timesteps.}
\end{center}
\end{table}

\begin{figure}[ht]
\begin{center}
\includegraphics[scale=0.80]{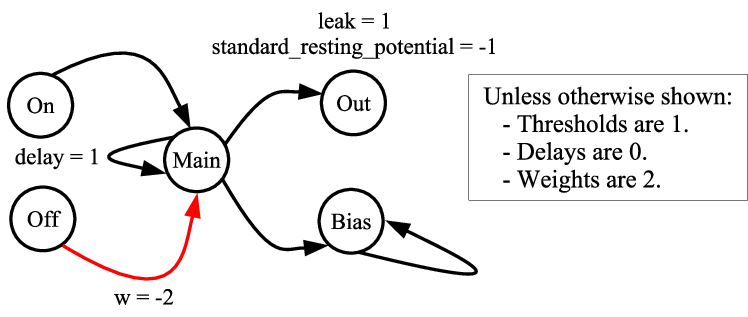}
\caption{\label{fig:network_4_more_leak} A network to demonstrate leak further.}
\end{center}
\end{figure}

To further illustrate leak, we next use the network in Figure~\ref{fig:network_4_more_leak}.
This network resets $Out's$ threshold to 1, and adds a delay of 1 to the synapse from
$Main$ back to itself.

When we start $Main$ spiking, the delay causes it to spike every other timestep rather than
every timestep.  As such, $Out$ leaks its charge to 0 then -1 every other timestep, at which
point the synapse from $Main$ increments it to 2.  As such, $Out$ 
never fires.  

\begin{table}[ht]
\begin{center}
\begin{tabular}{|l|c|ccccc|ccccc|l|}
\hline
Timestep & Apply & \multicolumn{5}{c|}{Fire} &
                   \multicolumn{5}{c|}{Charge at end of timestep} \\
         & spike to & $Main$ & $On$ & $Off$ & $Out$ & $Bias$ &
                       $Main$ & $On$ & $Off$ & $Out$ & $Bias$ \\
\hline
  0 & $Main$ & - & - & - & - & - & 16 & 0 & 0 & -1 & 0  \\
  1 &    -   & * & - & - & - & - & 0 & 0 & 0 & 1 & 2  \\
  2 &    -   & - & - & - & - & * & 2 & 0 & 0 & 0 & 2  \\
  3 &    -   & * & - & - & - & * & 0 & 0 & 0 & 1 & 4  \\
  4 &    -   & - & - & - & - & * & 2 & 0 & 0 & 0 & 2  \\
  5 &    -   & * & - & - & - & * & 0 & 0 & 0 & 1 & 4  \\
  6 &    -   & - & - & - & - & * & 2 & 0 & 0 & 0 & 2  \\
  7 &    -   & * & - & - & - & * & 0 & 0 & 0 & 1 & 4  \\
\hline
\end{tabular}
\caption{\label{tab:network_4_more_leak} Activity of the network in Figure~\ref{fig:network_4_more_leak}.  The leak causes $Out$ to to leak its charge down to -1 every other timestep, and it never fires.}
\end{center}
\end{table}

In Table~\ref{tab:network_4_more} we demonstrate that a neuron's charge cannot go
below its standard resting potential.  We use the same network as in Table~\ref{tab:network_4_more},
but apply the input spike to $Off$.  That causes $Off$ to fire at timestep 1, which sends -2
units of charge to $Main$.  This is shown as $Main's$ charge at the end of timestep 1.
It is reset to 0 at the beginning of timestep 2.

\begin{table}[ht]
\begin{center}
\begin{tabular}{|l|c|ccccc|ccccc|l|}
\hline
Timestep & Apply & \multicolumn{5}{c|}{Fire} &
                   \multicolumn{5}{c|}{Charge at end of timestep} \\
         & spike to & $Main$ & $On$ & $Off$ & $Out$ & $Bias$ &
                       $Main$ & $On$ & $Off$ & $Out$ & $Bias$ \\
\hline
  0 & $Off$  & - & - & - & - & - & 0 & 0 & 16 & -1 & 0  \\
  1 &    -   & - & - & * & - & - & -2 & 0 & 0 & -1 & 0  \\
  2 &    -   & - & - & - & - & - & 0 & 0 & 0 & -1 & 0  \\
\hline
\end{tabular}
\caption{\label{tab:network_4_more} Demonstrating that if charge goes below the standard resting potential, it is reset to the standard resting potential.}
\end{center}
\end{table}

\clearpage

\subsection{Absolute Refractory Period}

\begin{figure}[ht]
\begin{center}
\includegraphics[scale=0.80]{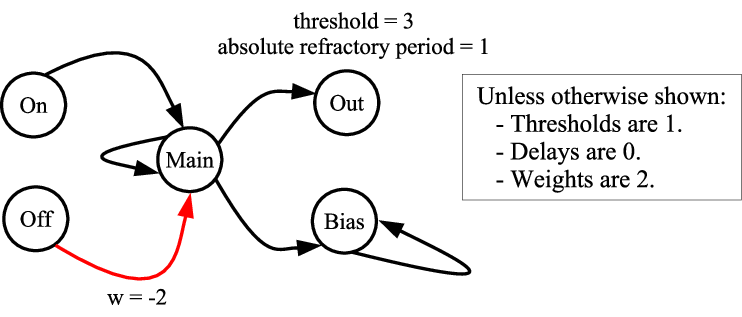}
\caption{\label{fig:network_5_abs_ref} A network to demonstrate the absolute refractory
period.}
\end{center}
\end{figure}

The network in Figure~\ref{fig:network_5_abs_ref} demonstrates the absolute refractory
period.  
This network configures $Out$ to so that its threshold is 2, and its 
absolute refractory period is 1.

We show the activity in Table~\ref{tab:network_5_abs_ref}.  We start $Main$ spiking, 
and because $Out's$ threshold is 3, it takes two cycles to have $Out's$ charge exceed its
threshold.  In the next timestep, $Out$ fires and enters its absolute refractory period.
As such, it ignores the spike from $Main$ and its charge remains 0.  On the next timestep,
the absolute refractory period is over, and it starts accumulating charge.  It continues in 
this manner, spiking every three timesteps.

\begin{table}[ht]
\begin{center}
\begin{tabular}{|l|c|ccccc|ccccc|l|}
\hline
Timestep & Apply & \multicolumn{5}{c|}{Fire} &
                   \multicolumn{5}{c|}{Charge at end of timestep} \\
         & spike to & $Main$ & $On$ & $Off$ & $Out$ & $Bias$ &
                       $Main$ & $On$ & $Off$ & $Out$ & $Bias$ \\
\hline
  0 & $Main$ & - & - & - & - & - & 16 & 0 & 0 & 0 & 0  \\
  1 &    -   & * & - & - & - & - & 2 & 0 & 0 & 2 & 2  \\
  2 &    -   & * & - & - & - & * & 2 & 0 & 0 & 4 & 4  \\
  3 &    -   & * & - & - & * & * & 2 & 0 & 0 & 0 & 4  \\
  4 &    -   & * & - & - & - & * & 2 & 0 & 0 & 2 & 4  \\
  5 &    -   & * & - & - & - & * & 2 & 0 & 0 & 4 & 4  \\
  6 &    -   & * & - & - & * & * & 2 & 0 & 0 & 0 & 4  \\
  7 &    -   & * & - & - & - & * & 2 & 0 & 0 & 2 & 4  \\
  8 &    -   & * & - & - & - & * & 2 & 0 & 0 & 4 & 4  \\
  9 &    -   & * & - & - & * & * & 2 & 0 & 0 & 0 & 4  \\
\hline
\end{tabular}
\caption{\label{tab:network_5_abs_ref} Activity of the network in Figure~\ref{fig:network_5_abs_ref}.  Because of its absolute refractory period of one timestep, $Out$ does not accumulate charge
from $Main$ in timesteps 3, 6 and 9.}
\end{center}
\end{table}

\clearpage 

\subsection{Relative Refractory Period and Refractory Resting Potential}

\begin{figure}[ht]
\begin{center}
\includegraphics[scale=0.80]{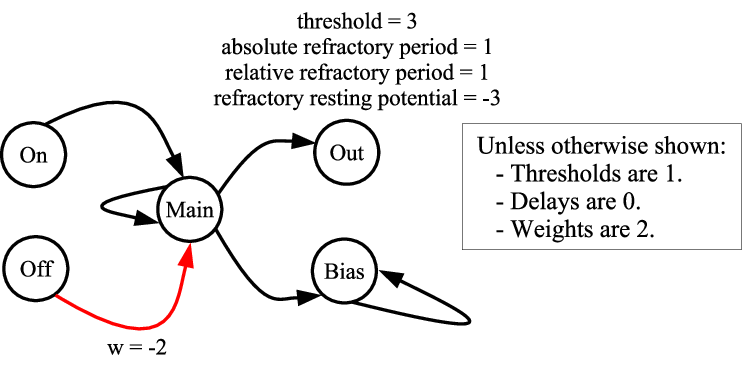}
\caption{\label{fig:network_6_rel_ref} A network to demonstrate the relative refractory
period and refractory resting potential.}j
\end{center}
\end{figure}

We use the network in Figure~\ref{fig:network_6_rel_ref} to demonstrate
the relative refractory period and refractory resting potential.  
As before, the threshold of~$Out$ is set to three.  Both refractory periods are set to 1, 
and the refractory resting potential is set to -3.  

When we start $Main$ spiking, $Main's$ synapse to $Out$ fires twice, causing $Out$ to
exceed its threshold at timestep 2.  It fires on timestep 3, setting its potential to
its refractory resting potential, and ignoring the spike from $Main$.  On timestep 4,
it enters its relative refractory period, and the spike from $Main$ increases its
charge to -1.  On timestep 5, the relative refractory period is over, so the charge is
set to the standard resting potential, which is zero.  It then receives its spike from
$Main$, which increases its charge to two.  On timestep 6, its charge exceeds its threshold,
so it fires on timestep 7, and repeats the process.
spikes to get $Out$ to fire again.  It repeats this pattern every four timesteps.

\begin{table}[ht]
\begin{center}
\begin{tabular}{|l|c|ccccc|ccccc|l|}
\hline
Timestep & Apply & \multicolumn{5}{c|}{Fire} &
                   \multicolumn{5}{c|}{Charge at end of timestep} \\
         & spike to & $Main$ & $On$ & $Off$ & $Out$ & $Bias$ &
                       $Main$ & $On$ & $Off$ & $Out$ & $Bias$ \\
\hline
  0 & $Main$ & - & - & - & - & - & 16 & 0 & 0 & 0 & 0  \\
  1 &    -   & * & - & - & - & - & 2 & 0 & 0 & 2 & 2  \\
  2 &    -   & * & - & - & - & * & 2 & 0 & 0 & 4 & 4  \\
  3 &    -   & * & - & - & * & * & 2 & 0 & 0 & -3 & 4  \\
  4 &    -   & * & - & - & - & * & 2 & 0 & 0 & -1 & 4  \\
  5 &    -   & * & - & - & - & * & 2 & 0 & 0 & 2 & 4  \\
  6 &    -   & * & - & - & - & * & 2 & 0 & 0 & 4 & 4  \\
  7 &    -   & * & - & - & * & * & 2 & 0 & 0 & -3 & 4  \\
  8 &    -   & * & - & - & - & * & 2 & 0 & 0 & -1 & 4  \\
  9 &    -   & * & - & - & - & * & 2 & 0 & 0 & 2 & 4  \\
 10 &    -   & * & - & - & - & * & 2 & 0 & 0 & 4 & 4  \\
 11 &    -   & * & - & - & * & * & 2 & 0 & 0 & -3 & 4  \\
\hline
\end{tabular}
\caption{\label{tab:network_6_rel_ref} Activity of the network in Figure~\ref{fig:network_6_rel_ref}.}
\end{center}
\end{table}

\clearpage 

\subsection{STDP - Potentiation}

\begin{figure}[ht]
\begin{center}
\includegraphics[scale=0.80]{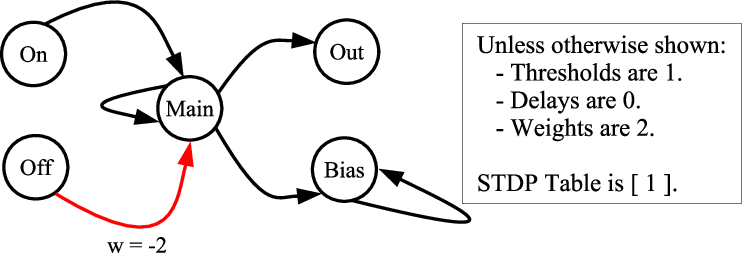}
\caption{\label{fig:network_7_stdp} A network to demonstrate simple STDP potentiation.}
\end{center}
\end{figure}

The network in Figure~\ref{fig:network_7_stdp} shows very simple potentiation.
The STDP table has just one element - one.  Whenever a synapse fires into a neuron,
and the neuron's charge exceeds the threshold at the end of the timestep, the synapse
will potentiate.

We apply a spike to $Main$ at timestep 0, which causes $Main$ to fire at timestep 1.
$Main's$ three synapses to $Out$, $Bias$ and $Main$ arrive during timestep 1, setting the
charge of all three neurons to two, which exceeds their thresholds.  As such, all three
synapses potentiate by one.  This is reflected in the charges of 3 in $Main$ and $Out$
in timestep 2.  $Bias'$ charge is 5, because the synapse from $Main$ has potentiated to 3,
and it received a charge of 2 from its self-synapse, which also potentiates. 

Those four synapses (three from $Main$ and one from $Bias$) keep potentiating until they
reach their maximum value of 7 (a Hardware Setting).

\begin{table}[ht]
\begin{center}
\begin{tabular}{|l|c|ccccc|ccccc|l|}
\hline
Timestep & Apply & \multicolumn{5}{c|}{Fire} &
                   \multicolumn{5}{c|}{Charge at end of timestep} \\
         & spike to & $Main$ & $On$ & $Off$ & $Out$ & $Bias$ &
                       $Main$ & $On$ & $Off$ & $Out$ & $Bias$ \\
\hline
  0 & $Main$ & - & - & - & - & - & 16 & 0 & 0 & 0 & 0  \\
  1 &    -   & * & - & - & - & - & 2 & 0 & 0 & 2 & 2  \\
  2 &    -   & * & - & - & * & * & 3 & 0 & 0 & 3 & 5  \\
  3 &    -   & * & - & - & * & * & 4 & 0 & 0 & 4 & 7  \\
  4 &    -   & * & - & - & * & * & 5 & 0 & 0 & 5 & 9  \\
  5 &    -   & * & - & - & * & * & 6 & 0 & 0 & 6 & 11  \\
  6 &    -   & * & - & - & * & * & 7 & 0 & 0 & 7 & 13  \\
  7 &    -   & * & - & - & * & * & 7 & 0 & 0 & 7 & 14  \\
\hline
\end{tabular}
\caption{\label{tab:network_7_stdp} Activity of the network in Figure~\ref{fig:network_7_stdp}.}
\end{center}
\end{table}

\begin{figure}[ht]
\begin{center}
\includegraphics[scale=0.80]{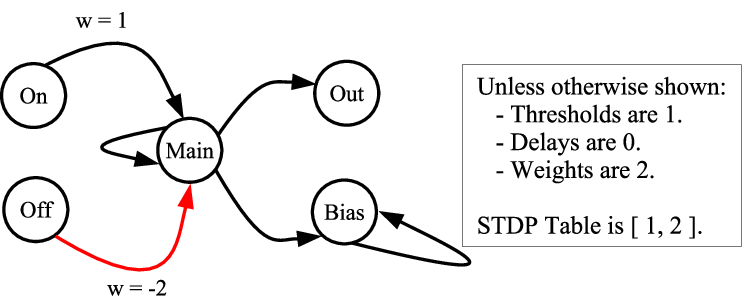}
\caption{\label{fig:network_8_stdp} A network to demonstrate a two-element STDP table.}
\end{center}
\end{figure}

In Figure~\ref{fig:network_8_stdp}, we make the potentiation table a little more complex.
The table size, $T$, equals 2, so indices 0 and 1 are used for potentiation.  There is no
depression.  We also reduce the weight of the synapse from $On$ to one.

For this example, we apply three spikes -- one to $On$ at timestep 0, one to $On$ at timestep 2,
and one to $Main$ at
timestep 2.  The spike from $On$ arrives at timestep 1, but is not enough to make $Main$
fire.  When we apply the spike to $Main$ at timestep 2, that causes $Main's$ charge to exceed
its threshold, so $Main$ will fire at timestep 3.  Accordingly, the synapse from $On$ to
$Main$ potentiates by one to have a weight of two.  At timestep 4, $Main$ receives two
spikes -- one from~$On$ with its new weight of two, and one from $Main$, also with a weight
of two.  This is shown by $Main's$ charge of 4 at timestep 4.  
The spike causes both synapses to potentiate by two to four.  

At timestep four we see that:

\begin{itemize}
\item $Main$ has a charge of 4, which has come from its self-synapse whose weight has potentiated to 4.
\item $Out$ has a charge of 4, which has come from its synapse from $Main$, whose weight has potentiated to 4.
\item $Bias$ has a charge of 4, which has come from its synapse from $Main$, whose weight has potentiated to 4, and its self-synapse, which has not yet potentiated.
\end{itemize}

\begin{table}[ht]
\begin{center}
\begin{tabular}{|l|c|ccccc|ccccc|l|}
\hline
Timestep & Apply & \multicolumn{5}{c|}{Fire} &
                   \multicolumn{5}{c|}{Charge at end of timestep} \\
         & spike to & $Main$ & $On$ & $Off$ & $Out$ & $Bias$ &
                       $Main$ & $On$ & $Off$ & $Out$ & $Bias$ \\
\hline
  0 & $On$   -   & - & - & - & - & - & 0 & 16 & 0 & 0 & 0  \\
  1 &    -   & - & * & - & - & - & 1 & 0 & 0 & 0 & 0  \\
  2 & $On$, $Main$ & - & - & - & - & - & 17 & 16 & 0 & 0 & 0  \\
  3 &    -   & * & * & - & - & - & 4 & 0 & 0 & 2 & 2  \\
  4 &    -   & * & - & - & * & * & 4 & 0 & 0 & 4 & 6  \\
\hline
\end{tabular}
\caption{\label{tab:network_8_stdp} Activity of the network in Figure~\ref{fig:network_8_stdp}.}
\end{center}
\end{table}

\clearpage 

\subsection{STDP - Depression}

\begin{figure}[ht]
\begin{center}
\includegraphics[scale=0.80]{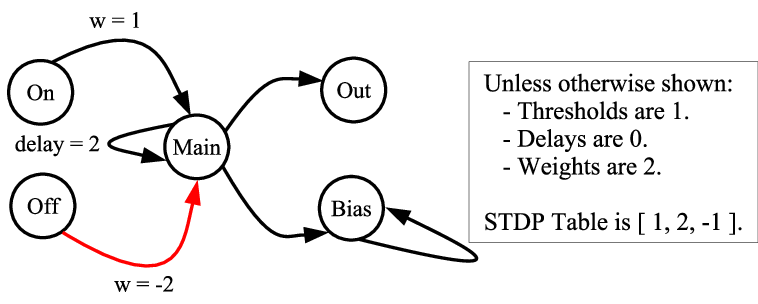}
\caption{\label{fig:network_9_stdp} A network to demonstrate simple STDP depression.}
\end{center}
\end{figure}

To demonstrate depression, we add another value to the STDP table, making it [ 1, 2, -1 ].
Since~$T = 3$, the first two elements of the table define potentiation, and the last defines
depression.  We use the network in Figure~\ref{fig:network_9_stdp} to demonstrate the depression.
This network is similar to the last network, except the self-synapse from $Main$ has a delay of
two, instead of zero.

To demonstrate depression, we apply spikes to both $Main$ and $On$ at timestep 0, and to $On$ at timestep 3.  The first two spikes cause $Main$ and $On$ to exceed their thresholds at timestep 0, and spike at timestep 1. The spike from $On$ arrives at 
$Main$ at timestep 1 and increments Main's potential to 1, which does not cause it to spike. Since $Main$ exceeded its potential at timestep 0, $On's$ synapse depresses to zero. 

The spike from $Main's$ self-synapse arrives at timestep 3, causing $Main$ to exceed its
threshold and spike at timestep 4.  The spike from $On$ arrives at timestep 4, and its
charge is zero.  That is why $Main$'s charge is zero at the end of timestep 4.  $On's$ synapse
at this point depresses to -1.

\begin{table}[ht]
\begin{center}
\begin{tabular}{|l|c|ccccc|ccccc|l|}
\hline
Timestep & Apply & \multicolumn{5}{c|}{Fire} &
                   \multicolumn{5}{c|}{Charge at end of timestep} \\
         & spike to & $Main$ & $On$ & $Off$ & $Out$ & $Bias$ &
                       $Main$ & $On$ & $Off$ & $Out$ & $Bias$ \\
\hline
  0 &    -   & - & - & - & - & - & 16 & 16 & 0 & 0 & 0  \\
  1 &    -   & * & * & - & - & - & 1 & 0 & 0 & 2 & 2  \\
  2 &    -   & - & - & - & * & * & 1 & 0 & 0 & 0 & 2  \\
  3 &    -   & - & - & - & - & * & 3 & 16 & 0 & 0 & 4  \\
  4 &    -   & * & * & - & - & * & 0 & 0 & 0 & 4 & 11  \\
\hline
\end{tabular}
\caption{\label{tab:network_9_stdp} Activity of the network in Figure~\ref{fig:network_9_stdp}.}
\end{center}
\end{table}

\begin{figure}[ht]
\begin{center}
\includegraphics[scale=0.80]{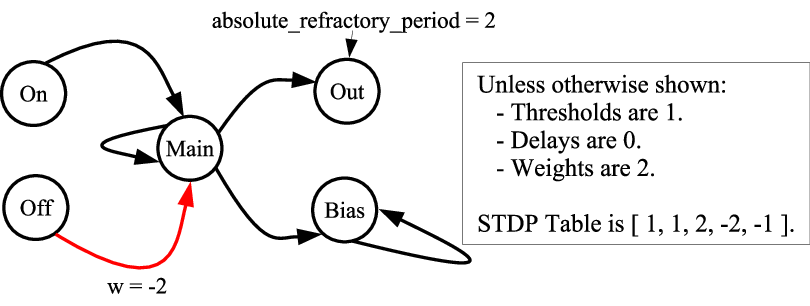}
\caption{\label{fig:network_a_stdp} A second network to demonstrate depression.}
\end{center}
\end{figure}

Depression can do things that may see odd.  Here's a last example where a synapse depresses
its weight despite the fact that its neuron is constantly firing.  The network is shown
in Figure~\ref{fig:network_a_stdp}.  The STDP table is larger: [ 1, 1, 2, -2, -1 ].
Since~$T = 5$, the first three elements define potentiation, and the last two define depression.

In the network, $Out$ has an absolute refractory period of 2.

We start demonstrating the activity by applying a spike to $Main$ at time 0.
This causes $Main$ to continually spike starting at timestep 1.  The synapse from $Main$
arrives at $Out$ in timestamp 1, causing it to spike at timestep 2.  This also causes the
synapse from $Main$ to potentiate its weight by two to four.  

In timesteps 2 and 3, $Out$ is in its absolute refractory period, which means the two spikes
from $Main$ are ignored.  Moreover, they cause the synapse depress, first by two and then by one,
so that its weight is now one.  In timestep 4, $Main$ goes back to standard operation and
receives a spike with weight 1 from main.  That is why its charge is one.  At timestep 5,
it receives another spike, raising its potential to 2.  Accordingly, it fires at timestep 6,
and the synapse from $Main$ potentiates to three.

In timesteps 2 and 3, $Out$ is in its absolute refractory period, which means the two spikes
from $Main$ are ignored, and the synapse depresses to 1 and then to zero.  $Out$ will never spike
again.

\begin{table}[ht]
\begin{center}
\begin{tabular}{|l|c|ccccc|ccccc|l|}
\hline
Timestep & Apply & \multicolumn{5}{c|}{Fire} &
                   \multicolumn{5}{c|}{Charge at end of timestep} \\
         & spike to & $Main$ & $On$ & $Off$ & $Out$ & $Bias$ &
                       $Main$ & $On$ & $Off$ & $Out$ & $Bias$ \\
\hline
  0 &    -   & - & - & - & - & - & 16 & 0 & 0 & 0 & 0  \\
  1 &    -   & * & - & - & - & - & 2 & 0 & 0 & 2 & 2  \\
  2 &    -   & * & - & - & * & * & 4 & 0 & 0 & 0 & 6  \\
  3 &    -   & * & - & - & - & * & 6 & 0 & 0 & 0 & 10  \\
  4 &    -   & * & - & - & - & * & 7 & 0 & 0 & 1 & 13  \\
  5 &    -   & * & - & - & - & * & 7 & 0 & 0 & 2 & 14  \\
  6 &    -   & * & - & - & * & * & 7 & 0 & 0 & 0 & 14  \\
  7 &    -   & * & - & - & - & * & 7 & 0 & 0 & 0 & 14  \\
  8 &    -   & * & - & - & - & * & 7 & 0 & 0 & 0 & 14  \\
  9 &    -   & * & - & - & - & * & 7 & 0 & 0 & 0 & 14  \\
 10 &    -   & * & - & - & - & * & 7 & 0 & 0 & 0 & 14  \\
\hline
\end{tabular}
\caption{\label{tab:network_a_stdp} Activity of the network in Figure~\ref{fig:network_a_stdp}.}
\end{center}
\end{table}

\clearpage 

\subsection{STDP - Synapses with Spikes ``In Flight''}

\begin{figure}[ht]
\begin{center}
\includegraphics[scale=0.80]{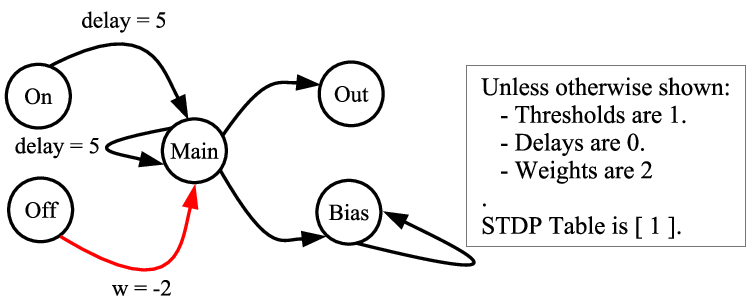}
\caption{\label{fig:network_c_flight} A network to show STDP on synapses with spikes in flightj.}
\end{center}
\end{figure}

For this we use a very simple network with a one-element STDP table.  It is shown
in Figure~\ref{fig:network_c_flight}.

When we run this, we'll apply spikes to $On$ at times 0, 2 and 3. That
fires the synapse from $On$ to $Main$ at times 1, 3 and 4, which means
that those synapses will fire into $Main$ at times 6, 8 and 9. When the
first of these arrives at $Main$, you'll note that the other two spikes
are ``in flight'' on the neuron.  However, when charge is addeed at
timestep 6, the synapse potentiates, so when the second spike arrives
at time 8, its weight is 3. Similarly, when the third spike arrives
at time 9, its weight is 4.

\begin{table}[ht]
\begin{center}
\begin{tabular}{|l|c|ccccc|ccccc|l|}
\hline
Timestep & Apply & \multicolumn{5}{c|}{Fire} &
                   \multicolumn{5}{c|}{Charge at end of timestep} \\
         & spike to & $Main$ & $On$ & $Off$ & $Out$ & $Bias$ &
                       $Main$ & $On$ & $Off$ & $Out$ & $Bias$ \\
\hline
  0 &    -   & - & - & - & - & - & 0 & 16 & 0 & 0 & 0  \\
  1 &    -   & - & * & - & - & - & 0 & 0 & 0 & 0 & 0  \\
  2 &    -   & - & - & - & - & - & 0 & 16 & 0 & 0 & 0  \\
  3 &    -   & - & * & - & - & - & 0 & 16 & 0 & 0 & 0  \\
  4 &    -   & - & * & - & - & - & 0 & 0 & 0 & 0 & 0  \\
  5 &    -   & - & - & - & - & - & 0 & 0 & 0 & 0 & 0  \\
  6 &    -   & - & - & - & - & - & 2 & 0 & 0 & 0 & 0  \\
  7 &    -   & * & - & - & - & - & 0 & 0 & 0 & 2 & 2  \\
  8 &    -   & - & - & - & * & * & 3 & 0 & 0 & 0 & 2  \\
  9 &    -   & * & - & - & - & * & 4 & 0 & 0 & 3 & 6  \\
\hline
\end{tabular}
\caption{\label{tab:network_c_flight} Activity of the network in Figure~\ref{fig:network_c_flight}.}
\end{center}
\end{table}

\clearpage
\section{Concluding Remarks}

All of the tables in this paper were created with the TENNLab software framework
and the RAVENS simulator~\cite{spp:21:sfc,psd:18:ten}.  The README for RAVENS in
the framework shows how to create these networks and process them to generate this
information.

\section{Acknowledgements}

This material is based on work supported by
the Air Force Research Laboratory, Information Directorate (FA8750-19-1-0025 and FA8750-21-1-1018).
The authors thank ChaoHui Zheng for writing the RAVENS simulator,
Bryson Gullett for writing the RAVENS-to-microcontroller compiler,
and Charlie Rizzo for hand-tooling networks.

\bibliographystyle{plain}
\bibliography{conference_101719}

\end{document}